# Title: A Foldable and Agile Soft Electromagnetic Robot for Multimodal Navigation in Confined and Unstructured Environments

## Author Information

### Affiliations

**State Key Laboratory of Fluid Power and Mechatronic Systems, Key Laboratory of Soft Machines and Smart Devices of Zhejiang Province, Center for X-mechanics, Department of Engineering Mechanics, Zhejiang University, Hangzhou 310027, People's Republic of China.**

Zhihao Lv[†], Xiaoyong Zhang[†], Mengfan Zhang, Xiaoyu Song, Xingyue Liu, Yide Liu, Shaoxing Qu, Guoyong Mao[*]

### Contributions

G.M. and Z.L. conceived the project. Z.L. and X.Z. fabricated and characterized the M-SEMR. Z.L. developed the theoretical and numerical model. X.Z. built dynamic simulations for the robot model. Z.L., G.M., and X.Z. designed the experiments. Z.L. and G.M. designed the figures and movies. M.Z. and X.S. assisted with the experiments. Z.L., X.Z., and G.M. analyzed the results and prepared and revised the manuscript. M.Z., X.S., X.L., Y.L., and S.Q. review & edit the paper. G.M. and S.Q. supervised the research.

[†]These authors contributed equally to this work: Zhihao Lv, Xiaoyong Zhang

### Corresponding author

[*]Correspondence to: Guoyong Mao
[*]Corresponding author. Email: guoyongmao@zju.edu.cn (G.M.)

### Abstract

Multimodal locomotion is crucial for an animal's adaptability in unstructured wild environments. Similarly, in the human gastrointestinal tract, characterized by viscoelastic mucus, complex rugae, and narrow sphincters like the cardia, multimodal locomotion is also essential for a small-scale soft robot to conduct tasks. Here, we introduce a small-scale compact, foldable, and robust soft electromagnetic robot (M-SEMR) with more than nine locomotion modes designed for such a scenario. Featuring a six-spoke elastomer body embedded with liquid metal channels and driven by Laplace forces under a static magnetic field, the M-SEMR is capable of rapid transitions (< 0.35 s) among different locomotion modes. It achieves exceptional agility, including high-speed rolling (818 mm/s, 26 BL/s), omnidirectional crawling, jumping, and swimming. Notably, the robot can fold to reduce its volume by 79%, enabling it to traverse confined spaces. We further validate its navigation capabilities on complex terrains, including discrete obstacles, viscoelastic gelatin surfaces, viscous fluids, and simulated biological tissues. This system offers a versatile strategy for developing high-mobility soft robots for future biomedical applications.

## Introduction

Multimodal locomotion enables organisms to adapt to complex environments, thereby enhancing their survival chances[1–7]. Inspired by these biological systems, researchers have developed various small-scale robots capable of multimodal locomotion to achieve functional versatility[8–16] with applications in path navigation[17,18], inspection and detection[19,20], and targeted drug delivery[21–23]. These capabilities are particularly critical for operation within confined and unstructured settings, such as the human gastrointestinal (GI) tract[24,25] which poses significant challenges: viscoelastic mucus, complex rugae, and narrow sphincters (e.g., the cardia) often constrain the mobility of existing soft robots. Current devices frequently suffer from limited adaptability to terrain irregularities, leading to functional failure during abrupt transitions or posture changes like flipping[9–11,15,17]. Therefore, developing a soft robot that combines a configurable structure with high mobility is essential for navigating such unstructured terrains[24]. Achieving this goal requires a synergy of material selection and structural design.

Previously, diverse smart materials, including shape memory alloys[10,26,27]/polymers[28], pneumatic systems[29,30], stimuli-responsive polymers[31–33,34], dielectric elastomers[35–37], and magnetic composites[23,8,14,38], have been extensively investigated. However, their widespread application in such environments is often hindered by intrinsic limitations, ranging from slow response speeds and bulky tethering systems to safety concerns regarding high voltage or difficulties in localized control. Soft electromagnetic actuators (SEMAs) driven by Laplace forces offer a compelling alternative, featuring millisecond-scale response, programmable deformations, and precise control[39–41].

Beyond materials, structural design is equally decisive. Crawling robots, typically employing bipedal or multi-legged structures, demonstrate excellent maneuverability. For instance, a bipedal robot developed by Mao et al. achieved a remarkable speed of 70 BL/s[11], while a bio-inspired multi-legged design from Ji et al. exhibited omnidirectional mobility[17]. However, neither design is suitable for performing in complex terrains. Swimming robots typically feature streamlined, elongated bodies for propulsion in aquatic environments[42], yet lack terrestrial mobility. Jumping robots, which utilize bistable mechanisms[43] or elastic energy storage systems[44] for rapid energy release, are effective for obstacle traversal but difficult for sustained locomotion. Rolling robots offer energy efficiency on flat surfaces, serving as a pathway to enhance mobility for robots, but typically struggle in unstructured environments and lack integration with other modes[30,32,34,36,37,45–47]. For instance, a humidity-responsive robot developed by Xu et al. achieved a slow rolling speed of 5.8 BL/s[32]. In contrast, a dielectric elastomer-based design by Li et al. reached only 0.98 BL/s on flat terrain[36]. While recent electromagnetic rollers achieved high speeds of 14 BL/s[48], they remain limited to only a single locomotion mode, failing to address the complex navigational needs of the unstructured environment.

Drawing inspiration from the Golden Wheel Spider and the Moroccan Flic-flac Spider, which utilize postural adjustments for rapid multimodal transitions between different modes[49,50] (Fig. S1A), we developed a compact, foldable, and multimodal soft electromagnetic robot (M-SEMR) designed for robust navigation in unstructured environments like the human GI tract (Fig. 1A). The M-SEMR features a six-spoke elastomer structure embedded with liquid metal channels, enhancing its ability to traverse undulating and viscous surfaces. Driven by Laplace forces within a global static magnetic field (such as an MRI environment), the robot is designed to fold significantly to traverse narrow constraints like the cardia, and subsequently unfold in the stomach to deliver drugs to target areas. By optimizing the actuation strategy and structure with a theoretical model, the M-SEMR can dynamically adjust among nine locomotion modes and postures (the most abundant modes among small-scale robots), thereby demonstrating greater locomotion versatility than existing small-scale soft robots with multimodal locomotion (Figs. 1B-C). Specifically, the M-SEMR utilizes two distinct postures, standing and lying, to enable modes, including rolling, crawling, walking, jumping, swimming, steering, backing, and posture switching, all achieved through control strategies rather than extra reconfigurable actuators. Notably, the M-SEMR can achieve a maximum rolling speed of 818 mm/s, complete a posture switch in 0.35 s, and perform in-place turning maneuvers. Furthermore, the robot demonstrated amphibious capabilities, including underwater deployment and underwater locomotion. The M-SEMR is also driven across various complex environments, including a gelatin gel surface, a discrete stepped terrain, viscous fluid, and a simulated gastric surface, providing a versatile solution for traversing viscoelastic mucus and complex rugae.

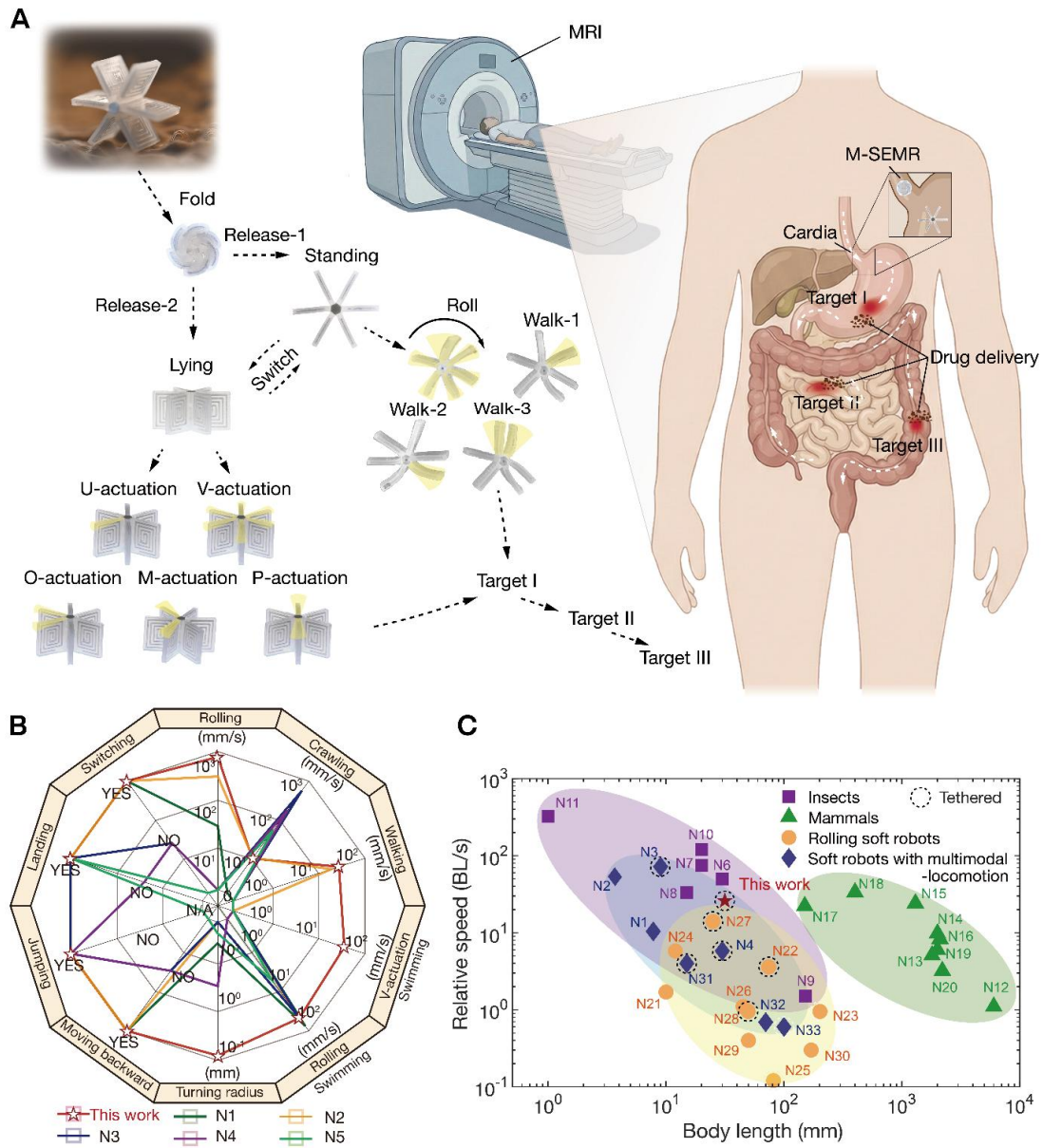

**Fig. 1. Multimodal locomotion and performance of the M-SEMR.** (**A**) The M-SEMR can be compactly folded into a cylindrical configuration to reduce its occupied volume. Upon release, it can switch actively between two postures. The standing posture supports rolling or walking in three modes. The lying posture supports U-actuation, V-actuation, or crawling with three dual-module coordinated actuation modes. (**B**) Comparison of locomotion modalities between the M-SEMR and other reported soft robots with multimodal locomotion. Details are provided in Supplementary Table S1. (**C**) Relationship between the maximum locomotion speed and body length of the M-SEMR, representative insects, rolling soft robots, and soft robots with multimodal locomotion. Details are provided in Supplementary Table S2.

# Results

## Design and fabrication

The M-SEMR is a fully soft and modular robotic system actuated in a static magnetic field, adopting a spoke configuration composed of six identical driven modules, which are soft electromagnetic actuators[39], made of a molded elastomeric shell embedded with liquid metal channels as soft conductive coils (Fig. 1A). Each module can be driven individually by Laplace forces resulting from a sequence of currents applied to the soft conductive coil. The modules are assembled using mortise and tenon joints, which minimize redundant components and overall system mass (Fig. SB-C). The entire M-SEMR weighs only 3.84 g and exhibits an internal resistance of 0.19 $\Omega$ per module.

To ensure the structural robustness of M-SEMR during its rolling, we characterize the mechanical response of the elastomer (Fig. S2A) and conduct a buckling analysis of a single module. Because the whole body of the rolling M-SEMR is periodically supported by a single module (Fig. S2B), this may result in a buckle that leads to motion destabilization. Considering the buckling analysis (Fig. S2C), flexibility, and fabrication of the module, the dimensions of the module are set to 16 mm × 14 mm × 2 mm (Fig. S2D). And the cross-section of the liquid metal channels is a square with dimensions of 0.5 mm × 0.4 mm. The gap between adjacent liquid metal channels is 1 mm to ensure electrical insulation and structural strength of a single module. The number of modules in an M-SEMR affects its rolling performance and terrain adaptability. Adding more modules to one M-SEMR improves the robot's peripheral envelope roundness, making it closer to an ideal circle (Fig. S3A), thereby enhancing the smoothness and continuity of rolling (Fig. S3B). Thus, too many modules reduce the foldability and obstacle clearance of the rolling M-SEMR. Here, we developed a quantitative model to balance these competing criteria, which led to the selection of a six-module design. (Fig. S3C). For actuation, the M-SEMR operates within a uniform magnetic field (Fig. S4) and is powered by a designed pulse width modulation (PWM) controller (Fig. S5). Each module can be independently actuated using square wave current with adjustable frequency and duty cycle (Fig. S6), enabling diverse deformation, including bending, twisting, compression, and extension (Fig. S7). Experiments demonstrate that such a design of M-SEMR supports fast multimodal locomotion and adaptability to complex terrains.

## Rolling

By maintaining a constant current and magnetic field direction while actuating the modules of the rolling M-SEMR sequentially, the resulting Laplace force vector is dynamically modulated to achieve targeted actuation across specific modules. Based on the dynamic analysis and actuation strategy optimization (supplementary materials), it is essential for fast rolling to shorten the startup time. A simple solution is to increase the current amplitude. As an example, when the pulse current increases from 1 A to 1.5 A. the startup time reduces from 40 ms to 20 ms (Fig. 2A), with a 30% decrease in the transition time between States 2 and 3 (Fig. 2B). The motion process of the rolling M-SEMR involves complex dynamic and finite deformations, making it challenging to describe with a simple theoretical model. Here, instead, we simplify the M-SEMR to be a fully rigid body and build a dynamic theoretical model to analyze the behavior of the M-SEMR during startup time. Details about this dynamic theoretical model are provided in the Supplementary Materials. From the theoretical prediction, the minimal startup time is 38 ms, higher than 20 ms from experiments (Fig. S11). From experimental observation in Fig. 2A, the soft module C2 first bends itself, so that the Laplace force contributes much more work in a short time than the prediction from the theoretical dynamic model. To better understand this, we conducted two simulations with an ideal rigid-body model and a semi-rigid model incorporating flexible hinges (Fig. S12A and Table S3). The rigid-body model yielded a startup time of 38 ms, matching the result from dynamic theory. The semi-rigid model predicted a startup time of 24.5 ms, close to the experimental result. Simulations comparing the two models confirmed that the flexible structure enables the actuator to perform an order of magnitude more work under an actuation duration of 20 ms (Fig. S12B), a result of the flexible structure. These findings highlight the importance of structural compliance in the design of M-SEMR.

To achieve rapid and continuous rolling of the M-SEMR, square-wave currents were sequentially applied to individual modules from C2 to C1 in an anticlockwise order (Movie S1). We defined the actuation signals into two types: global actuation signal and local actuation signal (Fig. S5B). The specific frequencies used for each type in different experiments are detailed in Table S4. Fig. S13 shows representative displacement-time, velocity-time, and corresponding acceleration-time curves of the M-SEMR subjected to various actuation frequencies of square wave currents. The rolling speed increases with actuation frequency and reaches a maximum of 818 mm/s (26 BL/s) at 60 Hz (Fig. 2C and Movie S1), representing the fastest rolling velocity reported for small-scale soft robots. However, the pulse duration of the 60-Hz square-wave current (the global actuation signal) is only 16.7 ms. This duration is shorter than the M-SEMR's startup time, making it unsuitable for the continuous rolling of the M-SEMR. As a result, the rolling success rate decreases sharply to 10% (Fig. 2D). To overcome this issue, we implemented a hybrid actuation strategy combining a short time of low-frequency actuation with a subsequent high-frequency actuation (Fig. S14). This approach substantially improved the rolling success rate to 80% (Fig. 2D), preserving a high rolling speed of 698 mm/s (see the data of 60P in Fig. S13D). A sequence of snapshots of the rolling M-SEMR is shown in Fig. 2E. During such rapid rolling, the robot exhibits intermittent jumping, which improves its adaptability to terrains with unstructured surfaces. Balancing the rolling speed and locomotion stability from Fig. 2D, a global actuation frequency of 40 Hz was selected for the following demonstration if there is no extra statement. with such a actuation frequency, the M-SEMR demonstrated robust rolling across a variety of substrates, including stainless steel, silicone, PMMA, polyurethane, sandpaper, and paper (Fig. 2F and Movie S1). On all tested surfaces, the robot achieved high speeds exceeding 500 mm/s and a maximum of 643 mm/s on paper. The trajectory analysis, as shown in the displacement-time curves of Fig. S15, demonstrates that M-SEMR achieves extra stable rolling on both polyurethane and PMMA surfaces with high consistency.

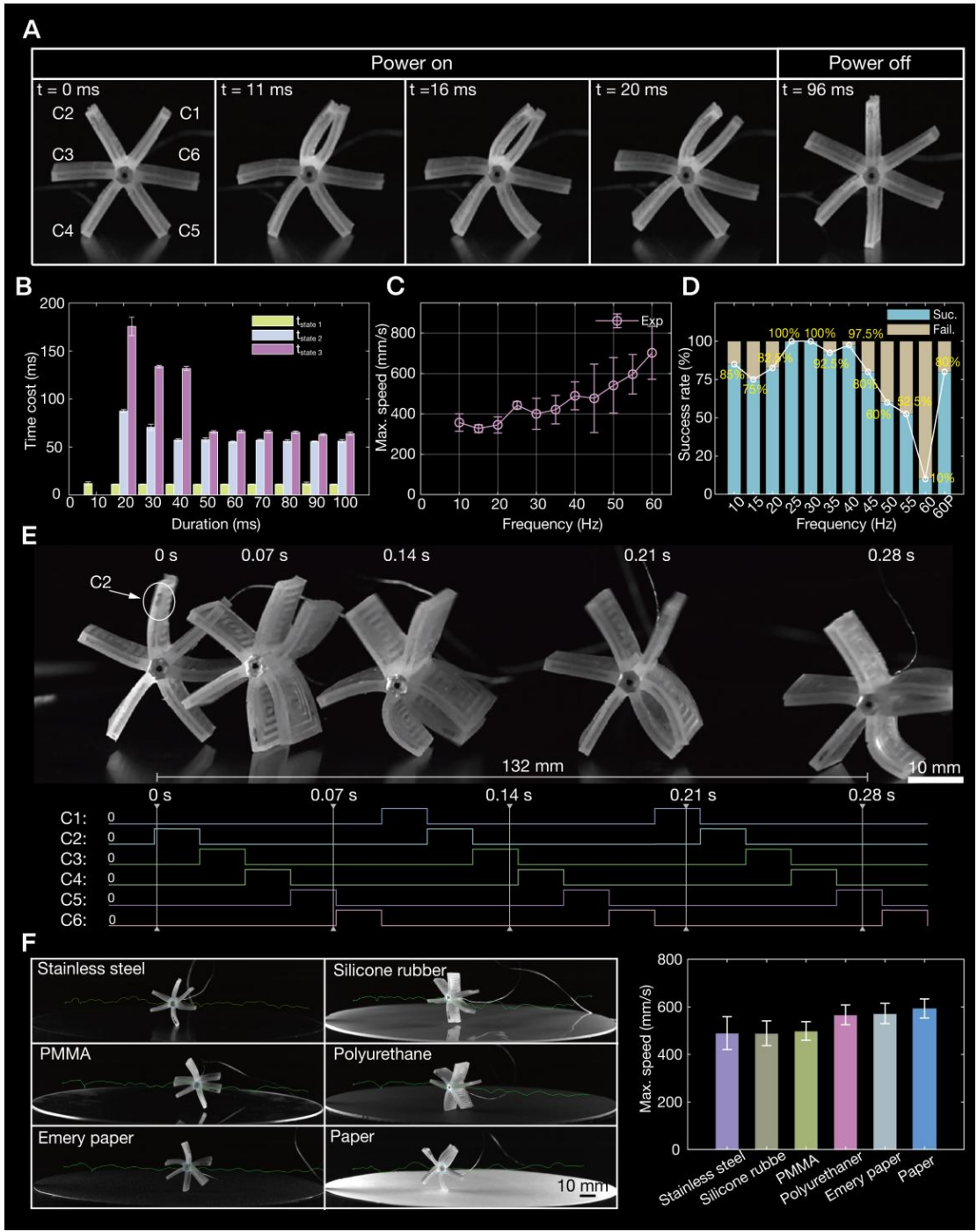

**Fig. 2. Performance of the rolling M-SEMR.** The amplitude of the current is 1.5 A. (**A**) Snapshots of the posture of M-SEMR subjected to a 20-ms pulse current. (**B**) Time costs for the M-SEMR to reach different states with various pulse durations. The robot does not reach state 2 with a 10-ms pulse current. (**C**) Maximum rolling velocity vs. actuation frequency of the M-SEMR subjected to a 1.5 A square wave current. (**D**) Success rate of the M-SEMR rolling from initial state to state 3 as a function of actuation frequency. There are 40 measurements conducted for each actuation frequency. (**E**) Snapshots of the rolling M-SEMR subjected to a sequence of 60 Hz square wave currents, followed by the global actuation signals. The peak speed is 818 mm/s (26 BL/s). (**F**) Trajectories (green lines) of the M-SEMR's mass center during continuous rolling on various surfaces, along with corresponding peak speeds illustrated on the right side. All error bars represent the standard deviation of five measurements.

## Walking and crawling

In addition to the rolling mode, the M-SEMR also exhibits three walking modes: walk-1, walk-2, and walk-3 (Fig. 3A-C, Movie S2). In walk-1, only one top module, such as C1 or C2, of the M-SEMR is actuated. As an example, in Fig. 3A, C1 bends back and forth under the Laplace force that induced the rotation of the whole body, producing an inchworm-like gait to the left side with an average speed of 16 mm/s (Fig. S16A). If only C2 is actuated, the M-SEMR will walk towards the opposite side so that the walking direction of the M-SEMR can be well controlled. In walk-2, only one bottom module is actuated, such as C5 in Fig. 3B. When C5 is driven by the Laplace force and bends up forward, C4 is the only support for the M-SEMR, which induces the bending of C4 and the forward movement of the whole M-SEMR. Once the Laplace force is removed, the M-SEMR tends to return to its standing posture (Fig. S16B). With such locomotion, the M-SEMR can walk at a speed of 24.26 mm/s. Similar to the Walk-1, the M-SEMR can switch its walk direction by switching the actuated bottom module. In Walk-3, the top two modules are actuated simultaneously using a pair of square wave currents, which have the same amplitude but a phase difference of half a period. The mechanism of this mode is similar to that of walk-1. In such a mode, the robot can walk at a speed of 23.76 mm/s (Fig. 3C, Fig. S16C), which is higher than that of walk-1.

The rolling and walking modes are based on the bending deformation of modules. The M-SEMR can also use the twist of modules to achieve crawling locomotion. Different from the standing posture of previous locomotion modes, crawling requires a lying posture. The relationship between the maximum torsion angle and the duration of a pulse current was experimentally characterized. As shown in Figure 3D, the torsion angle increases with the pulse current duration and reaches a maximum of 41° once the duration exceeds 10 ms. Accordingly, a 10-ms actuation pulse was chosen to ensure rapid and sufficient deformation for fast straight movement. A 4-ms pulse was chosen for a slight deformation, which is beneficial for M-SEMR to turn in place. As Fig. 1A U-actuation shows, when only a module is actuated, the M-SEMR can move a short distance. Selective activation of modules C1 through C6 enables omnidirectional crawling, and the robot retains mobility even with partial actuator failure, demonstrating strong fault-tolerant capabilities.

To demonstrate the maneuverability of the M-SEMR in its lying posture, we developed three dual-module actuation strategies, termed O-, M-, and P-actuation (inspired by the positional isomers of xylene, Fig. 3E-G), which depend on the positions of activated modules. The fundamental principle in all strategies is the polarization difference between the two activated modules. When the two modules are driven with an opposite polarization, a combined force propels the robot forward or backward in a linear path. By selectively activating different pairs of modules using these three strategies, the M-SEMR can achieve six-direction crawling. When the two modules are driven with an identical polarization, it creates a torque that causes the robot to rotate clockwise or anticlockwise (Fig. S17). We implemented these two pairs of driving currents with three dual-module actuations. O-actuation (Fig. 3E): Two adjacent modules (such as C2 and C3, 60° apart) are activated. The angle between their driving forces is 120° (opposite polarization) or 60° (identical polarization), resulting in a slow linear forward crawling speed of 1.84 mm/s and a minimum turning radius of 6.4 mm (Movie S3). M-actuation (Fig. 3F): Two modules separated by 120° (such as C2 and C4) are activated. This changes the angle between the forces to 60° (opposite polarization) or 120° (identical polarization), yielding a higher speed of 5.21 mm/s and a smaller turning radius of 3.7 mm (Movie S3). P-actuation (Fig. 3G): Two opposing modules (180° apart, such as C2 and C5) are activated. The force angles become 0° (opposite polarization) or 180° (identical polarization), enabling the highest speed of 10.51 mm/s and in-place rotation (Movie S3). This capability was conclusively demonstrated by controlling the M-SEMR to precisely trace a complex "ZJU" trajectory, which required seamless integration of forward/backward motions and clockwise/anticlockwise rotations (Fig. 3H and Movie S3).

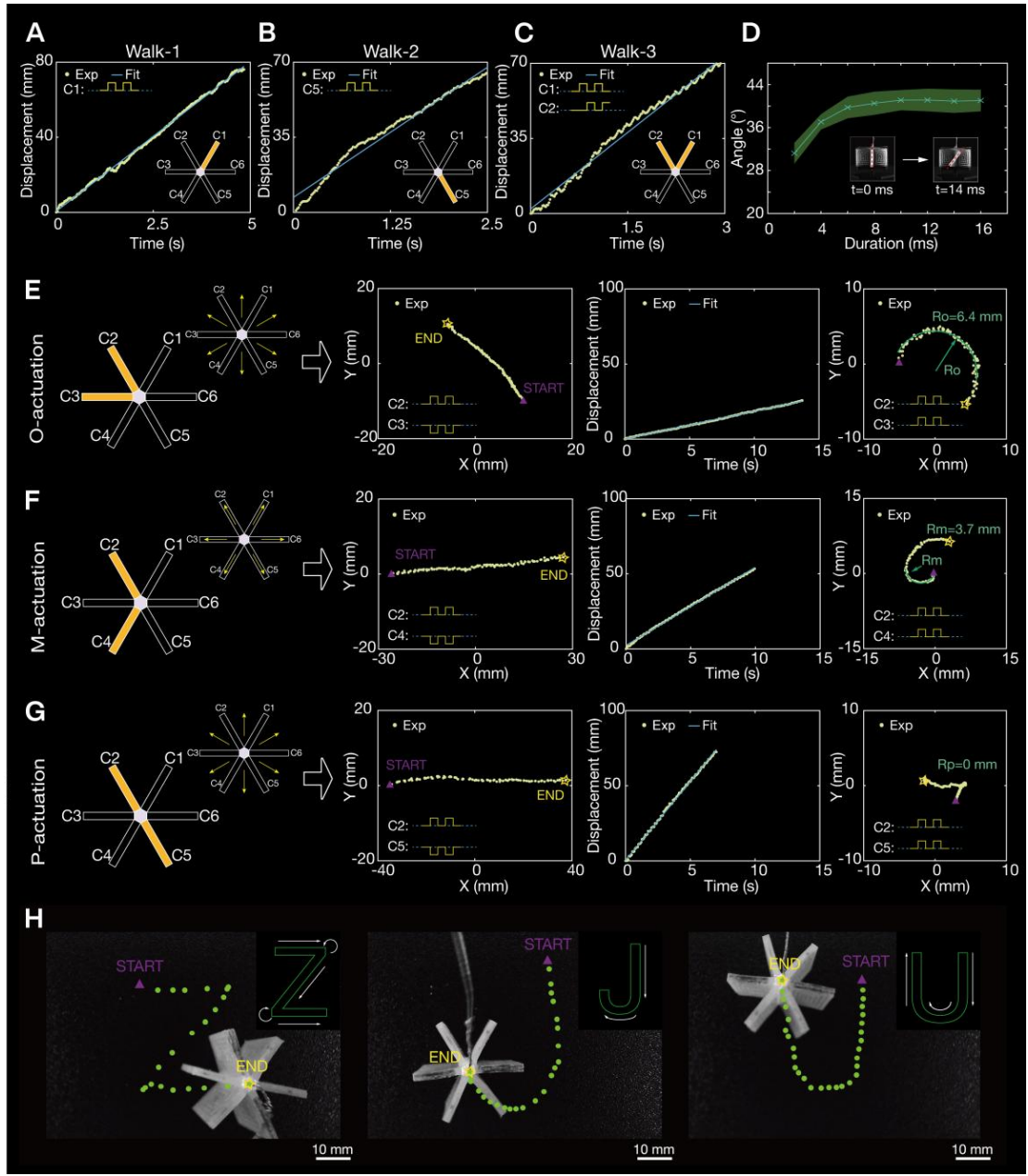

**Fig. 3. Characterization of walking and crawling M-SEMR.** The driven current in this panel is 1.5 A. Time-displacement curves of M-SEMR adopting three walking strategies: walk-1 (**A**), walk-2 (**B**), and walk-3 (**C**) in a standing posture with highlighted modules actuated (bottom right corner). The slope of the linear fit corresponds to the average speed. (**D**) Dynamic response of the module torsion angle to single pulse-width variation in the M-SEMR lying posture. Each data point is obtained from 20 measurements. Three dual-module coordinated actuation strategies: O-actuation (**E**), M-actuation (**F**), and P-actuation (**G**) in the M-SEMR lying posture. Trajectories during linear motion, time-displacement curves of linear movement, and turning trajectories for the M-SEMR under different driving strategies. Yellow-highlighted

modules indicate the current actuation modules. Ro, Rm, and Rp represent the turning radii corresponding to their respective actuation strategies. (**H**) The M-SEMR completes the "ZJU" trajectory using the P-actuation strategy.

## Posture switching

In previous sections, we have introduced nine different types of locomotion modes of M-SEMR, which can be divided into two groups according to the corresponding postures, lying and standing. To freely switch among all nine locomotion modes, it is important to find a control strategy that enables the M-SEMR to change its posture easily (Fig. 4A). From experimental observation, the rolling M-SEMR enters an unstable oscillatory state under high-frequency actuation. When the current is cut off in such an unstable state, the M-SEMR naturally falls into the lying posture by gravity in 350 ms (Fig. S18 and Movie S4). From experiments, we also observe that the M-SEMR can switch from the lying to the standing posture through a sequence of different driven currents. During this process of standing up, the torque from the Laplace force plays an important role, which overcomes the gravitational torque, elevating the center of mass (CoM) of the M-SEMR. A detailed analysis shows that each actuation module (C1-C6) sequentially receives a 50-ms pulse. Initial activation of C1 establishes a ground contact at C5 and C6, forming a lever mechanism with the distal end serving as the pivot. Subsequent activation of C2 dynamically shifts the contact points to C6 and C1. After 300 ms, the robot re-establishes the standing posture (Fig. S19). Compared with the transition from standing to lying, the switching from lying to standing involves more complex dynamics. Kinematic decoupling identifies two primary components:(1) sequential rotational actuation of individual modules and (2) elevation of the mass center through coordinated body rotation about distal contact points. The first component, induced by signal transitions, contributes minimally to net mechanical work and can be neglected. The second component is modeled as a module transitioning from vertical to horizontal, allowing magneto-mechanical analysis of torque balance and energy transfer (Fig. S20A-D). Analytical results demonstrate that this rotational elevation reduces instantaneous power by elongating the movement path of the mass center while maintaining constant total work, thereby facilitating an energy-efficient transition (Fig. S20E-F). In Fig. 4B, we plot the transition time from lying posture to standing posture as a function of actuation frequency, indicating that the time cost increases as the frequency exceeds 20Hz. Additionally, it is difficult for the M-SEMR to switch from lying to standing posture at frequencies above 60 Hz (such as 70 Hz).

To demonstrate the advantages of posture switching of the M-SEMR, we design a task for the M-SEMR to navigate in a constrained environment (Fig. 4C-H and Movie S4) in 46 s, utilizing two locomotion modes and two posture-switching actions. A rectangular obstacle was set, leaving a passageway with a height of 26 mm, to obstruct the locomotion of the M-SEMR toward the right side. It needs to be pointed out that the height of the M-SEMR is 27 mm in the static state, and the maximum height of the rolling M-SEMR is 31.5 mm (Fig. 4C). Considering this, the M-SEMR cannot cross through the 26-mm passageway by rolling. Instead, the M-SEMR switched the posture from standing to lying in 0.5 s, actuated at 67 Hz, resulting in the height of the M-SEMR changed to 16 mm (lower than passageway) (Fig. 4D). Once the transition was complete, the M-SEMR switched to the U-actuation mode and crawled through the passageway in 29 s, actuated at 16.7 Hz (Figs. 4E-F). Upon reaching the right side, the M-SEMR switched to its standing posture from a lying posture at a global actuation frequency of 20 Hz (Fig. 4G). Then, subjected to a 50-ms pulse, the M-SEMR started a new rolling process (Fig. 4H). A significant portion of the 46s task time is due to the manual switching of control commands. This duration could be substantially reduced by implementing automated control. Through this task, the M-SEMR demonstrated its adaptability in constrained environments, thereby providing a robust approach for navigation in height-restricted terrains.

The robustness of robots, encompassing both fault-tolerant and fail-safe strategies, is essential for their survival. The M-SEMR presented here exhibits remarkable robustness by integrating soft materials with a simple structure. As demonstrated in Fig. 3, the M-SEMR can maintain locomotion even if only a single module is functional, exhibiting exceptional fault tolerance. Experimentally, we also confirmed its excellent resilience to physical damage; the prototype was compressed to 6 mm (20% of its height) by a load and fully recovered its functionality after the load was removed (Fig. 4I, Fig. S21, and Movie S4). To further demonstrate its capabilities, we tested its locomotion under sustained compression in a narrow space where standard movement was impossible (Fig. 4J, Movie S4). Here, we designed two signals to selectively activate module pairs (C2 & C4; C1 & C5), inducing periodic bending that propels the robot forward (Fig. S22).

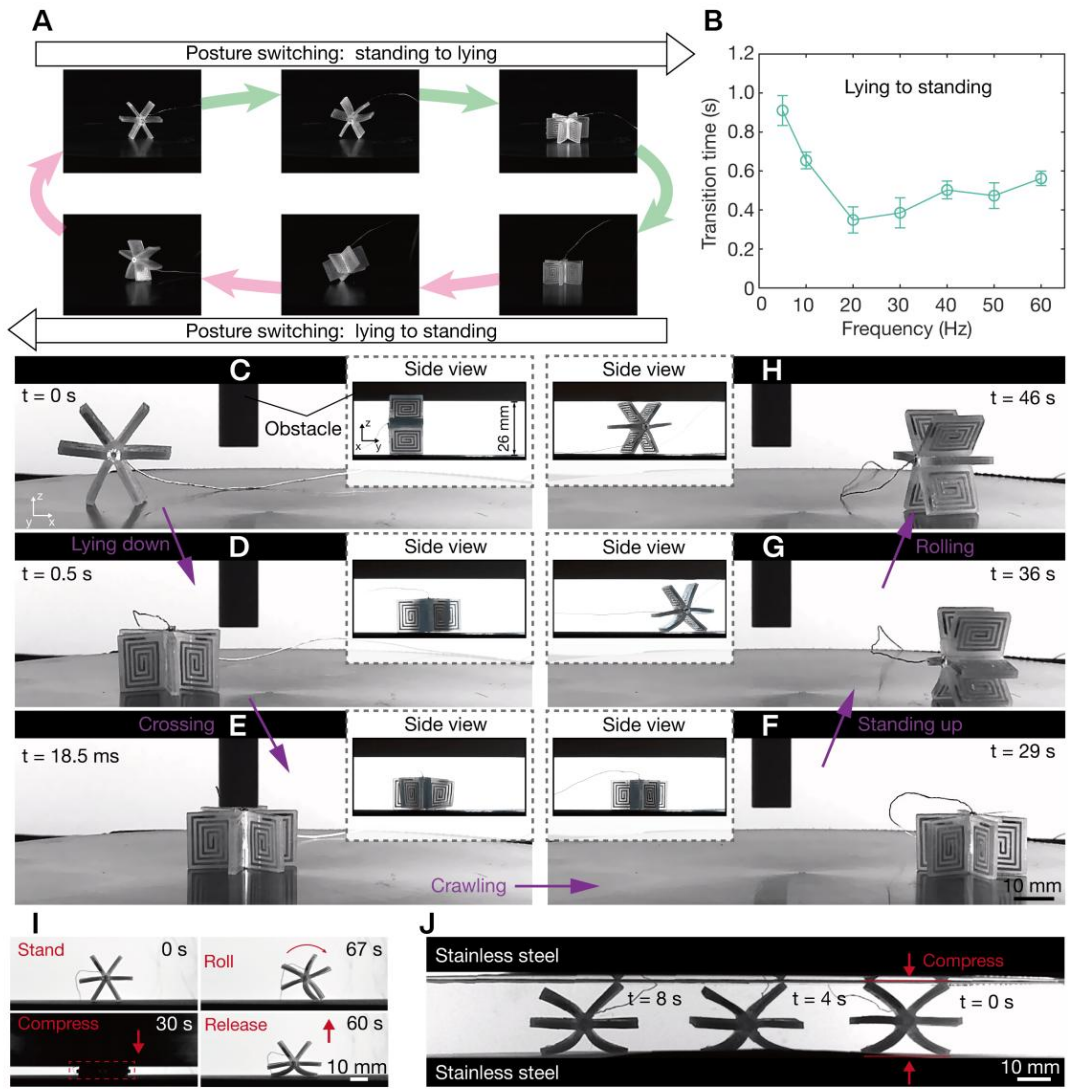

**Fig. 4. Posture switching and navigating through a constrained space of the M-SEMR.** (**A**) Active posture switching of the M-SEMR in steps. (**B**) Relationship between transition time and actuation frequency during M-SEMR's posture switching from lying to standing posture. (**C**) An obstacle is placed above the M-SEMR's forward path. (**D**) The M-SEMR switches to the lying posture. (**E**) The M-SEMR crawls through a narrow passageway. (**F**) The M-SEMR crawls a designated distance away from the passageway. (**G**) The M-SEMR switches back to the standing posture. (**H**) The M-SEMR continues to roll. (**I**) Robustness test showing the resilience of M-SEMR upon external loading. The robot resumes rolling after being pressed and fully flattened. (**J**) Locomotion of the M-SEMR under compression in a narrow space.

## Underwater locomotion

In addition to navigating on land, the M-SEMR can also realize underwater locomotion. Here, considering the future practical application of the M-SEMR in the gastrointestinal tract, we also demonstrated a new deployment approach, as shown in Fig. 5A, where the M-SEMR (31.5 mm in diameter) is folded into a compact cylindrical configuration (14.5 mm in diameter) and packaged within a dissolvable polyvinyl alcohol (PVA) film. It is noted that the M-SEMR's occupied volume is reduced to 21% in its folded state compared to its unfolded state. This approach enables the M-SEMR to cross a narrow opening, such as the gastric cardia, and then unfold autonomously in a water environment due to the dissolution of the PVA film. Fig. 5B and Movie S5 demonstrate the experimental deployment process of the folded M-SEMR to water through a tube (18 mm in inner diameter), including releasing, dissolving, and locomoting. The dissolving of the PVA film induced the release of the stored elastic energy in the M-SEMR, which is unfolded automatically. Subsequently, subjected to a 5 Hz square wave current, the unfolded M-SEMR performed a rolling locomotion toward the right side.

The water environment brings two new major factors in the locomotion of the M-SEMR, including the additional buoyancy force and the high drag force of water, which induce new experimental phenomena. Experimentally, we found that the M-SEMR can actively switch between its two underwater locomotion modes: rolling and V-actuation (Fig. 1, Fig. 5C, and Movie S5). In these two modes, all modules of the M-SEMR are driven in turns with a square wave current. The locomotion modes are highly frequency-dependent. For a low global actuation frequency, 5 Hz or 10 Hz, the M-SEMR can only locomote by rolling (Fig. S23). It is difficult for the M-SEMR to roll underwater at frequencies higher than 10 Hz, such as 15Hz. If the M-SEMR is in a lying posture, it can automatically switch to a standing posture and continue to roll. For a high global frequency, such as 40 Hz, the M-SEMR can only locomote with V-actuation. If the M-SEMR is in a standing posture, it can automatically switch to a lying posture and continue to locomote with V-actuation. We also tested the M-SEMR with higher global actuation frequencies, from 50 Hz to 100 Hz with a step size of 10 Hz, which shows the same phenomenon (Fig. S24 and Movie S5). We further quantified the performance of these modes. The M-SEMR's posture transition time in water is reduced to 200 ms, 150 ms faster than that on land (Fig. S23A). From our experiments, the M-SEMR achieves a maximum average rolling speed of 51.7 mm/s at a global frequency of 5 Hz (Fig. S23C-D) and an average locomotion speed of 35.1 mm/s with V-actuation at 50 Hz (Fig. S24).

To demonstrate the potential applications of the M-SEMR, we conducted two experiments as shown in Fig. S25. Here, we designed a controllable drug release module (Fig. S25A) that was integrated with the M-SEMR. This module consists of a chamber, platinum electrodes, and a cover with a microchannel in its center. The diameter of the microchannel is set to be 0.5 mm, which is small enough to seal the liquid drug in the chamber. In actuation, the electrodes are connected to a DC power supply for the electrolysis of the water in the chamber. Along with the electrolysis, oxygen and hydrogen accumulate around the electrodes, pushing the liquid drug out of the chamber through the microchannel in the cover. As shown in Fig. S25B and Movie S6, upon crawling to the target area, the M-SEMR releases its liquid drug by activating the drug module with a 10V voltage for 3 seconds. Following the release, the M-SEMR is controlled to mix the liquid drug with water. From the mechanism of the drug-release module, the drug delivery process can be well controlled for multiple releases. As evidenced in Fig. S25C and Movie S6, the M-SEMR crawled to three positions and released drugs with various dosages, demonstrating its controllable release capability.

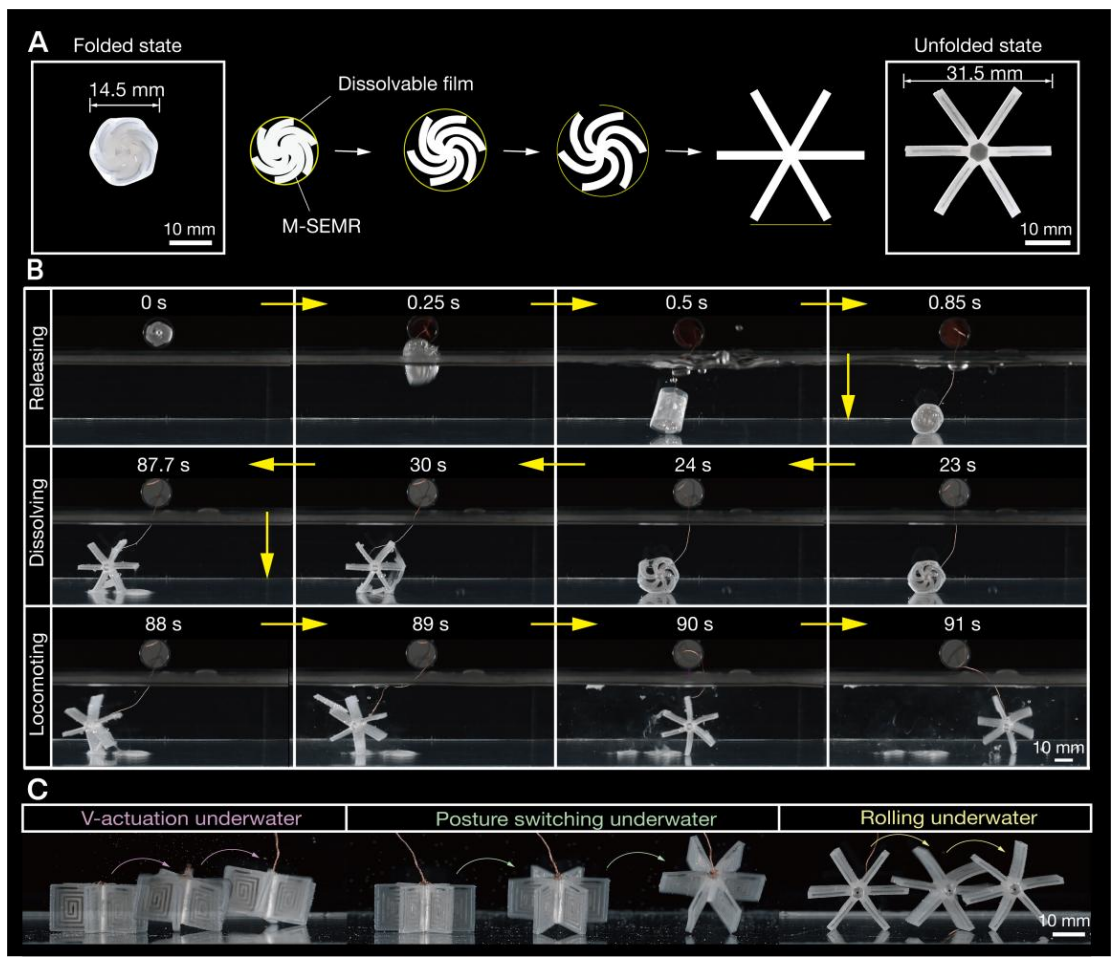

**Fig. 5. Release, locomotion, and function of an M-SEMR in water.** (**A**) Schematic illustration of the M-SEMR's folding, encapsulation, and autonomous release. The folded M-SEMR is encapsulated with a dissolvable film. (**B**) Snapshots of the folded M-SEMR at three stages: release, dissolution of the encapsulating film, and rolling locomotion. (**C**) locomotion with V-actuation, posture switching, and rolling of the M-SEMR.

## Off-road locomotion of M-SEMR

Most soft robots are operated in constrained and unstructured terrains, where environmental variations significantly impact their locomotion performance. The M-SEMR, however, leveraging its actuation mechanism and structural design, demonstrates environmental adaptability across diverse terrains.

To evaluate the M-SEMR's environmental adaptability, we conducted a series of locomotion experiments on various terrains (Movie S7). Results show that the M-SEMR can cross a series of small wedge blocks with a base length of 20 mm and varying inclinations from 6° to 38°, in less than 0.6 s (Fig. S26). Additionally, the M-SEMR can cross a series of small rectangular obstacles with heights ranging from 1 mm to 10 mm. This maximum height corresponds to approximately 0.37 times the M-SEMR's body height (Fig. S27A). Besides, we built two periodic surfaces, a square shape and a sinusoidal shape, on which the M-SEMR can locomote easily, as demonstrated in Figs. 6A, S27B, and S28. Additionally, we prepared a 3D-printed stomach with a surface topography resembling the gastric rugae, where the M-SEMR easily traversed in only 0.5 s, achieving a speed comparable to that on flat terrain (Fig. S29).

Further, we created a soft and adhesive surface made of a gelatin gel, on which the M-SEMR achieved an average speed of 226.2 mm/s at 20 Hz. This high speed is attributed to the M-SEMR's spoke structure, which minimizes the contact area during rolling and thereby effectively prevents adhesion to the gel surface (Fig. 6B). And from our experience, most small-scale robots cannot even walk on such a viscus surface [11]. Then, we tested the locomotion of the M-SEMR in a viscous non-Newtonian fluid made of yogurt solution (Fig. S30). The viscosity of the yogurt solution is 800 mPa·s at 1 rad/s, which is higher than that of gastric mucus, about 85 mPa·s[51]. In this viscous fluid, the M-SEMR, driven by a series of currents with a global frequency of 2 Hz, achieved a rolling speed of 34.5 mm/s. We also observed that the rolling locomotion was hindered at higher frequencies (such as 10 Hz) because of the high startup time of the rolling M-SEMR in viscous fluid (Fig. 6C). Apart from a solid surface or a liquid environment, the M-SEMR can roll from an aquatic to a terrestrial environment within 2 s, demonstrating its environmental adaptability (Fig. 6D).

Based on these results, we further demonstrated the M-SEMR's performance in diverse complex environments with varying structures and materials to comprehensively evaluate its potential in applications (Movie S8). For instance, we tested the M-SEMR within a 3D stomach model filled with viscous fluid to mimic the gastric environment. In this setting, the initially folded M-SEMR unfolded upon contact with fluid, navigated to a target area, and subsequently released the liquid drug in a controlled manner (Fig. 6E). These results demonstrated the robot's capability to function effectively in complex environments.

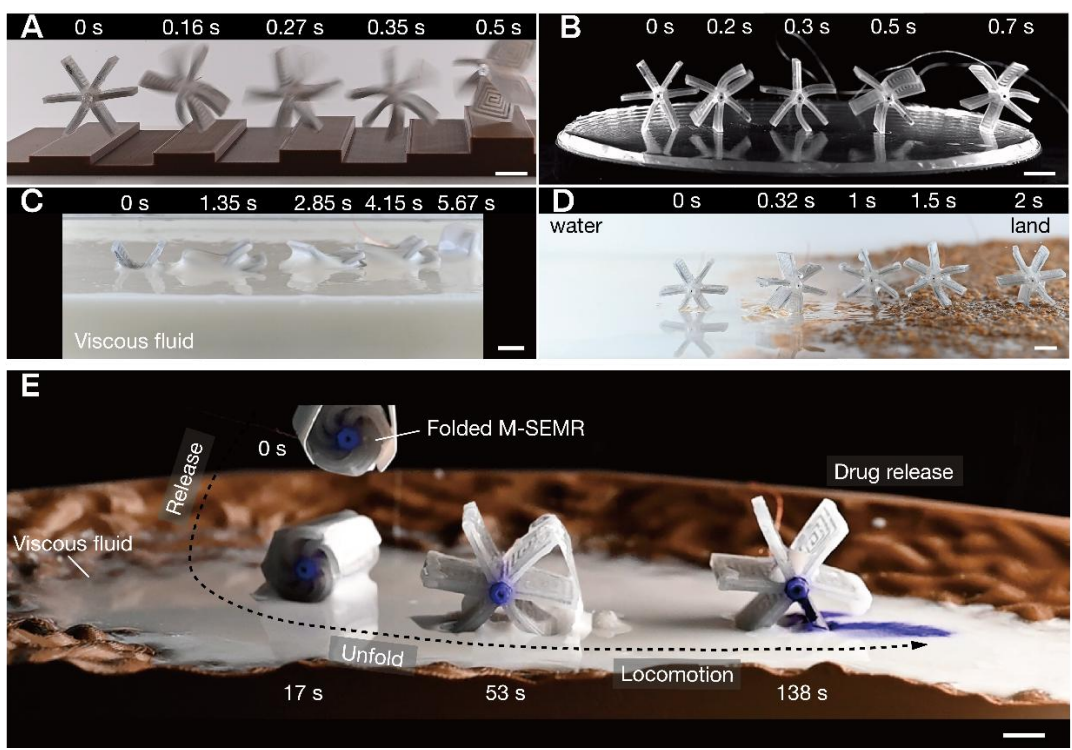

**Fig. 6. Off-road locomotion of M-SEMR.** (**A**) Snapshot of the M-SEMR locomoting on a periodic square-shaped surface. (**B**) Snapshot of the M-SEMR locomoting on a viscoelastic gelatin gel surface. (**C**) Snapshot of the M-SEMR locomoting in a viscous fluid. (**D**) Snapshot of the M-SEMR locomoting from water to land. (**E**) Snapshot of the M-SEMR functioning in a 3D-printed stomach. Scale bar is 10 mm.

## Discussion

In this work, we present a a small-scale compact, foldable and robust soft electromagnetic robot with nine locomotion modes, which owns the most abundant modes among small-scale robots, actuated exclusively by Laplace forces within a static global magnetic field generated by electromagnet (such as an MRI machine, which can greatly improve the mechanical efficiency of the M-SEMR[39]) and constructed using a modular multi-module design. The developed strategy in posture switching, between lying and standing in less than 0.5 s, enables the M-SEMR to change its locomotion modes among rolling, three walks, three crawls, and locomotion with U, V-actuation freely. In the standing posture, the M-SEMR achieves a top rolling speed of 818 mm/s (26 BL/s), representing the fastest reported rolling performance among small-scale soft robots. It also maintains a speed of up to 500 mm/s across various substrate surfaces, including stainless steel, silicone, PMMA, polyurethane, sandpaper, and paper.

Notably, the robot incorporates a foldable design that reduces its occupied volume to 21% of its original size, allowing it to access confined environments and subsequently deploy for task execution. The M-SEMR also shows high robustness in the experiments, enabling it to maintain locomotion with only a few modules and exhibiting resilience to physical damage, such as a high degree of compression. Experiments reveal that the M-SEMR with multimodal locomotion can cross various complex terrains, including discrete obstacles, gelatin-based tissue surfaces, viscous fluids, and simulated internal surface of the stomach, highlighting its potential in biomedical diagnostics and search-and-rescue operations. Moreover, we conducted an experiment where the robot locomoted in a 3D-printed stomach model and released drugs at a targeted position, highlighting its feasibility in a simulated gastric environment.

Despite these advancements, the current system faces several technical challenges. First, the molding method and handmade fabrication limit further miniaturization, which can be solved with advanced technology such as high-precision multi-material 3D printing. Second, the current M-SEMR is still a tethered version, and a custom design of micro-power and controller still needs to be developed. Third, even though we have demonstrated a series of locomotion modes and control strategies for the M-SEMR, we believe there is still room to explore. Fourth, although this study demonstrates enhanced terrain adaptability, the dynamical response of the M-SEMR in actuation requires further systematic investigation for better performance.

Crucially, the M-SEMR achieves multifunctionality without compromising the performance of individual functions, allowing a single robot to be repurposed for distinct tasks without the need for replacement[24]. This versatility is particularly advantageous for biomedical scenarios, positioning the demonstrated M-SEMR as a promising solution for future in vivo applications, ranging from targeted drug delivery and minimally invasive surgery to health monitoring.

## Methods

### Materials

The M-SEMR's architecture integrates six modular modules, each consisting of geometrically patterned soft elastomeric shells encapsulating liquid metal. The liquid metal, Galinstan (Dongguan Huatai Metal Material Technology Co., Ltd.), consists of gallium, indium, and tin with a mass ratio of 68.5:21.5:10. The elastomers were prepared by mixing Ecoflex-30 (Smooth-On Inc) and PDMS (Sylgard 184, Dow Corning) solution at a mass ratio of 10:1. A silicone adhesive (Sil-Poxy, Smooth-On) was used to bond the elastomers. The central support column in M-SEMR was fabricated from PLA (PLA, Bambu Lab). Additionally, a drug-release module was integrated into the robot system. This module was fabricated using 3D printing with a photopolymerizable resin (Rigid 10k resin, Formlabs) and features embedded inert platinum electrodes for water electrolysis.

### Fabrication

Step 1, mixing: The Ecoflex-30 solution was first prepared by mixing Part A and Part B with a mass ratio of 1:1. Separately, the PDMS solution was formulated with a 10:1 mass ratio of base monomer to curing agent. These two solutions were then thoroughly mixed at a mass ratio of 10:1 (Ecoflex-30: PDMS) using a planetary centrifugal mixer (BHZ-300, Beihong) at 2000 rpm for 3 minutes. Step 2, molding: The resulting mixture was poured into a PETG mold fabricated via a 3D printer (X1 Carbon, Bambu Lab). The mold filled with solution was placed in a vacuum chamber (100 mbar) for degassing for 3 minutes. An acrylic plate, which is fixed with clips, was used to remove excess material and cover the mold. Step 3, curing: The mold was placed in an oven at 50°C and cured for 25 minutes to form the elastomeric substrate. Step 4, demolding: The cured elastomers were carefully separated from the mold using tweezers, yielding two types of components: Elastomer 1 (with predefined microchannels) and Elastomer 2 (with a flat surface). Step 5, forming sealed modules: A silicone adhesive (Sil-Poxy, Smooth-On) was applied to the flat side of Elastomer 2, which was then aligned and bonded to Elastomer 1. The assembly was subsequently placed in an oven at 60°C for 3 minutes to complete the bonding process, resulting in individual and sealed elastomeric modules with empty microchannels. Step 6, liquid metal injection and wiring: For each module, liquid metal (Galinstan) was injected into its microchannels using a syringe with a 30G needle. Electrical wires were then connected to the injected liquid metal to form a functional actuation module. This was repeated for all six modules. Step 7, final assembly: The six functional modules were assembled into a multi-layer structure. The silicone adhesive was applied to the bonding interface of one module, which was then aligned and attached to its adjacent module. This bonding process was repeated, and the entire assembly was cured at 60°C to form the final six-layer elastomeric structure. Finally, the central support column, a 3D-printed PLA hexagonal prismatic support, was embedded at the center of the structure during the final assembly steps, completing the fabrication of the M-SEMR. More details can be found in the supplementary materials (Fig. S1).

## Control and power systems

The control system includes a microcontroller (Uno Rev3, Arduino) and a 16-channel PWM controller (PCA9685, Adafruit), which provides twelve independently programmable PWM signals for this setup. The drive system consists of six H-bridge motor driver modules (L298N, ST), each powered by a dedicated, independently regulated DC power supply (IT6833A, ITECH; or UDP6720, UNI-T). Every H-bridge module, controlled by two PWM signals, can output a programmable bidirectional current with a magnitude of up to 2 A.

## Numerical calculation and simulation

A custom Python script is used to optimize the module's thickness of the M-SEMR, along with the calculation of Laplace force, torque, and work during a single rolling cycle. We also developed MATLAB scripts for theoretical modeling to predict the startup time of a single rolling cycle, and the mechanical modeling of the rectangular driving module (Fig. S11). Results of both fully rigid and rigid-flexible coupled dynamic models were obtained using the software Adams (Fig. S12). Details of the simulation procedure are provided in the supplementary materials.

Trajectory tracking and speed measurement

The locomotion performances of the M-SEMR were recorded using different cameras for various experiments. (1) For rolling performance tests, a high-speed camera (ACS-1, NAC Image Technology) was used, recording at 2000 fps. The central black marker on the M-SEMR was tracked frame-by-frame using the software MVIEW (NAC Image Technology). A custom MATLAB script is used to calculate the instantaneous velocity considering the displacement across 10 frames (corresponding to a 10 ms interval) around the time point. (2) Walking and crawling motions were recorded using a high-speed camera (ACS-1) and a miniature camera module (DH-4k-003, Shenzhen Donghai Home Technology Co., Ltd.), respectively, both at 59.9 fps. The M-SEMR's position was extracted using the software Tracker (Version 6.1.2, Open Source Physics). The velocity was calculated as the average speed over a 33-ms interval around each recorded time point. (3) Underwater locomotion was recorded at 59.9 fps using a mirrorless camera (Z5, Nikon). Velocity was processed using the same method as depicted in (2). (4) Off-road locomotion in complex environments was recorded at 59.9 fps. Two cameras, the high-speed camera (ACS-1) and the mirrorless camera (Z5, Nikon), were used simultaneously to capture from different perspectives. Velocity analysis was performed following the method described in (2). In Figs. 3, S13, S15, S23, S24, and S31, we plotted displacement and velocity as functions of time of M-SEMR's locomotion in diverse environments.

## Acknowledgements

We thank Prof. Wei Yang, Prof. Tiefeng Li, and Prof. Pei Zhao for fruitful discussions. The authors also thank Dr. Qiyang Li for his technical assistance in the rheological experiment using the Discovery HR-20 analyzer (TA Instruments, USA). We use generative AI and AI-assisted discussion technologies in the writing process to improve the readability and language of the work.

Funding:

National Key R&D Program of China under Grants Nos. 2023YFB4704700

National Natural Science Foundation of China under grant Nos. 12372167, 12321002

111 Project of China (No. B21034)

## Ethics declarations

Competing interests

The authors declare that they have no competing interests.

## Data Availability

All data needed to evaluate the conclusions of the paper are available in the paper or the Supplementary Materials. Additional data related to this paper may be requested from the authors.